# Fast and Accurate Image Super Resolution by Deep CNN with Skip Connection and Network in Network

Jin Yamanaka[1], Shigesumi Kuwashima[1] and Takio Kurita[2][000-0003-3982-6750]

[1] ViewPLUS Inc., Chiyoda-ku Tokyo, 102-0084, Japan
[2] Hiroshima University, Hiroshima, 739-8527, Japan
{jin, kuwashima}@viewplus.co.jp, tkurita@hiroshima-u.ac.jp

**Abstract.** We propose a highly efficient and faster Single Image Super-Resolution (SISR) model with Deep Convolutional neural networks (Deep CNN). Deep CNN have recently shown that they have a significant reconstruction performance on single-image super-resolution. The current trend is using deeper CNN layers to improve performance. However, deep models demand larger computation resources and are not suitable for network edge devices like mobile, tablet and IoT devices. Our model achieves state-of-the-art reconstruction performance with at least 10 times lower calculation cost by Deep CNN with Residual Net, Skip Connection and Network in Network (DCSCN). A combination of Deep CNNs and Skip connection layers are used as a feature extractor for image features on both local and global areas. Parallelized 1x1 CNNs, like the one called Network in Network, are also used for image reconstruction. That structure reduces the dimensions of the previous layer's output for faster computation with less information loss, and make it possible to process original images directly. Also we optimize the number of layers and filters of each CNN to significantly reduce the calculation cost. Thus, the proposed algorithm not only achieves state-of-the-art performance but also achieves faster and more efficient computation. Code is available at https://github.com/jiny2001/dcscn-super-resolution.

## 1 Introduction

Single Image Super-Resolution (SISR) was mainly used for specific fields like security video surveillance and medical imaging. But now SISR is widely needed in TV, video playing, and websites as display resolutions are getting higher and higher while source contents remain between twice and eight times lower resolution when compared to recent displays. In other cases, network bandwidth is generally limited while the display's resolution is rather high. Recent Deep-Learning based methods (especially with deeply and fully convolutional networks) have achieved high performance in the problem of SISR from low resolution (LR) images to high resolution (HR) images. We believe this is because deep learning can progressively grasp both local and global structures on the image at same time by cascading CNNs and nonlinear layers. However, with regards to power consumption and real-time processing, deeply and fully convolutional networks require large computation and a lengthy processing time. In this paper, we propose a lighter network by optimizing the network structure with recent deep-learning techniques, as shown in Figure 1. For example, recent state-of-the-art deep-learning based SISR models which we will introduce at section 2 have 20 to 30 CNN layers, while our proposed model (DCSCN) needs only 11 layers and the total computations of CNN filters are 10 to 100 times smaller than the others.

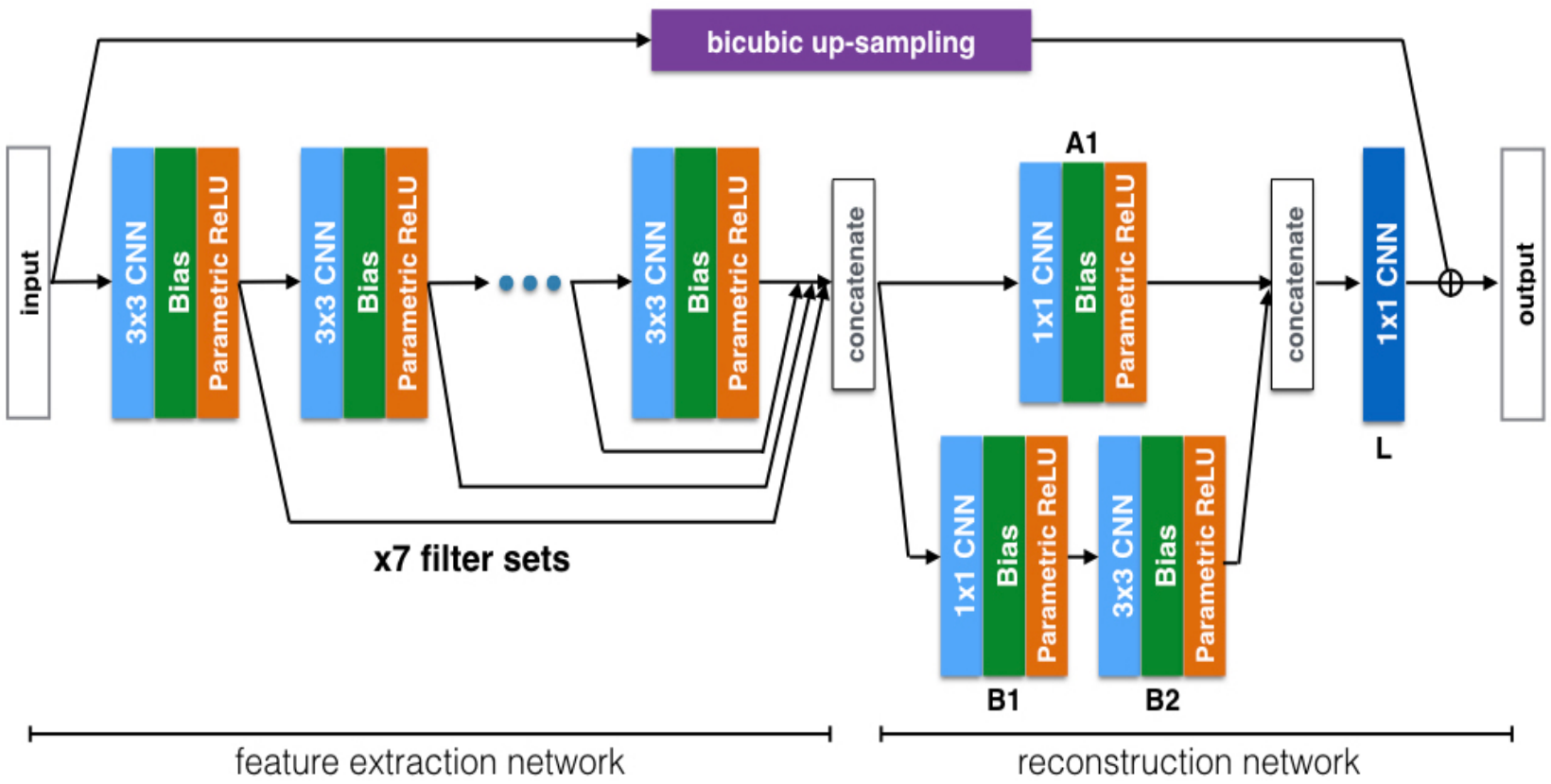

**Fig. 1.** Our model (DCSCN) structure. The last CNN (dark blue) outputs the channels of the square of scale factor. Then it will be reshaped to a HR image.

**Feature Extraction** In the previous Deep Learning-based methods, an up-sampled image was often used as their input. In these models, the SISR networks can be pixel-wise and its implementation becomes easier. However, they have 20-30 CNN layers in total and heavy computation is required for each up-sampled pixel. Furthermore, extracting features of up-sampled pixel is redundant, especially in the case of a scale factor of 3 or more. We use an original image as an input of our model so that the network can grasp the features efficiently. We also optimize the number of filters of each CNN layer and send those features directly to the image reconstruction network via skip connections.

**Image Detail Reconstruction** In the case of data up-sampling, the transposed convolutional layer (also known as a deconvolution layer) proposed by Matthew D. Zeiler [1] is typically used. The transposed convolutional layer can learn up-sampling kernels, however, the process is similar to the usual convolutional layer and the reconstruction ability is limited. To obtain a better reconstruction performance, the transposed convolutional layers need to be stacked deeply, which means the process needs heavy computation. So we propose a parallelized CNN structure like the Network in Network [2], which usually consists of one (or more) 1x1 CNN(s). Remarkably, the 1x1 CNN layer not only reduces the dimensions of the previous layer for faster computation with less information loss, but also adds more nonlinearity to enhance the potential representation of the network. With this structure, we can significantly reduce the number of CNN or transposed CNN filters. 1x1 CNN has 9 times less computation than 3x3 CNN, so our reconstruction network is much lighter than other deep-learning based methods.

## 2 Related Work

Deep Learning-based methods are currently active and showing significant performances on SISR tasks. Super-Resolution Convolutional Neural Network (SRCNN) [3] is the method proposed at this very early stage. C. Dong et al. use 2 to 4 CNN layers to prove that the learned CNN layers model performs well on SISR tasks. The authors concluded that using a larger CNN filter size is better than using deeper CNN layers. SRCNN is followed by Deeply-Recursive Convolutional Network for Image Super-Resolution (DRCN) [4]. DRCN uses deep (a total of 20) CNN layers, which means the model has huge parameters. However, they share each CNN's weight to reduce the number of parameters to train, meaning they succeed in training the deep CNN network and achieving significant performances.

The other Deep Learning-based method, VDSR [5], is proposed by the same authors of DRCN. VDSR uses Deep Residual Learning [6], which was developed by researchers from Microsoft Research and is famous for receiving first place in ILSVRC 2015 (a large image classification competition). By using residual-learning and gradient clipping, VDSR proposed a way of significantly speeding up the training step. Very deep Residual Encoder-Decoder Networks (RED) [7] are also based on residual-learning. RED contains symmetric convolutional (encoder) and deconvolutional (decoder) layers. It also has skip connections and connects instead to every two or three layers. Using this symmetric structure, they can train very deep (30 of) layers and achieve state-of-the-art performance. These studies therefore reflect the trend of "the Deeper the Better".

On the other hand, Yaniv Romano et al. proposed Rapid and Accurate Image Super Resolution (RAISR) [8], which is a shallow and faster learning-based method. It classifies input image patches according to the patch's angle, strength and coherence and then learn maps from LR image to HR image among the clustered patches. C. Dong et al. also proposed FSRCNN [9] as a faster version of their SRCNN [3]. FSRCNN uses transposed CNN to process the input image directly. RAISR and FRSCNN's processing speeds are 10 to 100 times faster than other state-of-the-art Deep Learning-based methods. However, their performance is not as high as other deeply convolutional methods, like DRCN, VDSR or RED.

# 3 Proposed Method

We started building our model from scratch. Started from only 1 CNN layer with small dataset and then grow the number of layers, filters and the data. When it stopped improving performance, we tried to change the model structure and tried lots of deep learning technics like mini-batch, dropout, batch normalization, regularizations, initializations, optimizers and activators to learn the meanings of using each structures and technics. Finally, we carefully chose structures and hyper parameters which will suit for SISR task and build our final model.

## 3.1 Model Overview

Our model(DCSCN) is a fully convolutional neural network. As shown in Figure 1, DCSCN consists of a feature extraction network and a reconstruction network. We cascade a set of CNN weights, biases and non-linear layers to the input. Then, to extract both the local and the global image features, all outputs of the hidden layers are connected to the reconstruction network as Skip Connection. After concatenating all of the features, parallelized CNNs (Network in Network [2]) are used to reconstruct the image details. The last CNN layer outputs the 4ch (or the channels of square of scale factor) image and finally the up-sampled original image is estimated by adding these outputs to the up-sampled image constructed by bicubic interpolation. Thus the proposed CNN model focusses on learning the residuals between the bicubic interpolation of the LR image and the HR original image.

In the previous studies, an up-sampled image was often used as their input for the Deep Learning-based architecture. In these models, the SISR networks will be pixel-wise. However, 20-30 CNN layers are necessary for each up-sampled pixel and heavy computation (up to 4x, 9x and 16x) is required, as shown in Figure 2. It also seems inefficient to extract a feature from an up-sampled image rather than from the original image, even from the perspective of the reconstruction process.

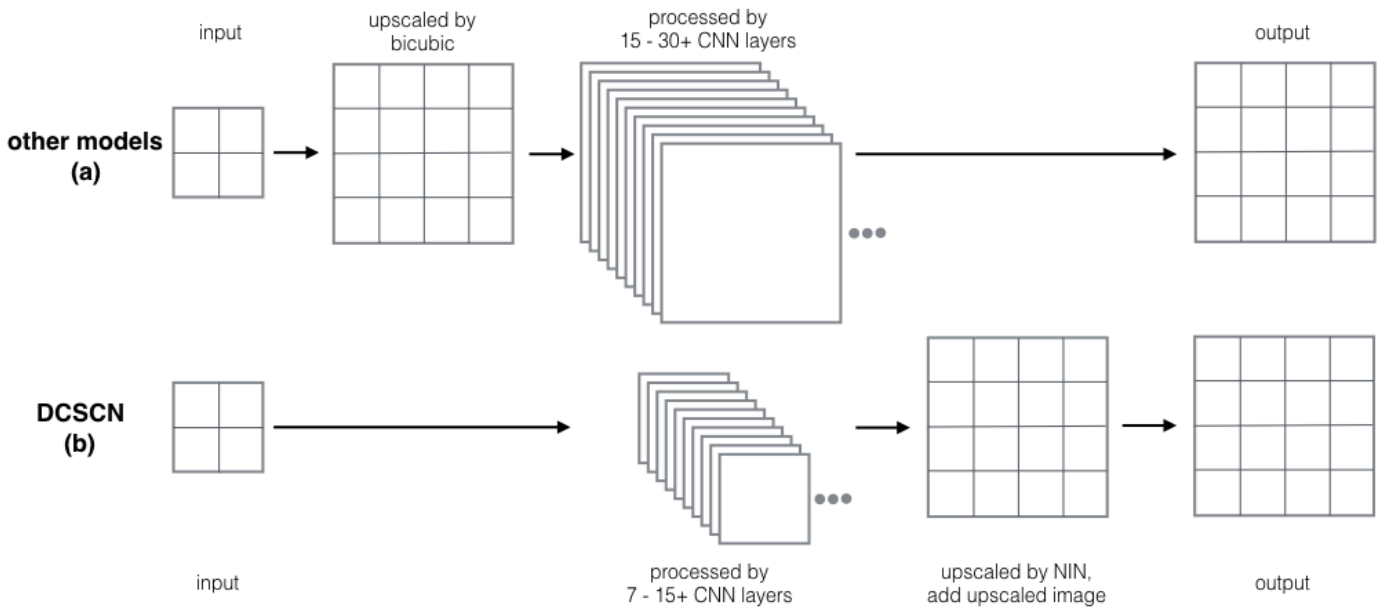

**Fig. 2.** Simplified process structures of (a) other models and (b) our model (DCSCN).

### 3.2 Feature Extraction Network

In the first feature extraction network, we cascade 7 sets of 3x3 CNN, bias and Parametric ReLU units. Each output of the units is passed to the next unit and simultaneously skipped to the reconstruction network. Unlike with other major deep-learning based large-scale image recognition models, the number of units of CNN layers are decreased from 96 to 32, as shown in Table 1. As discussed in Yang et al. [10], for model pruning, it is important to use an appropriate number of training parameters to optimize the network. Since the local feature is more important than the global feature in SISR problems, we reduce the features by the following layer and it results in better performance with faster computation. We also use the Parametric ReLU units as activation units to handle the "dying ReLU" problem [11]. This prevents weights from learning a large negative bias term and can lead to a slightly better performance.

### 3.3 Image Reconstruction Network

As stated in the Model Overview, DCSCN directly processes original images so that it can extract features efficiently. The final HR image is reconstructed in the last half of the model and the network structure is like in the Network in Network [2]. Because of all of the features are concatenated at the input layer of the reconstruction network, the dimension of input data is rather large. So we use 1x1 CNNs to reduce the input dimension before generating the HR pixels.

The last CNN, represented by the dark blue color in Figure 1, outputs 4 channels (when the scale factor s = 2) and each channel represents each corner-pixel of the up-sampled pixel. DCSCN reshapes the 4ch LR image to an HR(4x) image and then finally it is added to the bi-cubic up-sampled original input image. As with typical Residual learning networks, the model is made to focus on learning residual output and this greatly helps learning performances, even in cases of shallow (less than 7 layers) models.

**Table 1.** The numbers of filters of each CNN layer of our proposed model

| | Feature extraction network | | | | | | | Reconstruction network | | | |
|---|---|---|---|---|---|---|---|---|---|---|---|
| | 1 | 2 | 3 | 4 | 5 | 6 | 7 | A1 | B1 | B2 | L |
| DCSCN | 96 | 76 | 65 | 55 | 47 | 39 | 32 | 64 | 32 | 32 | 4 |
| c-DCSCN | 32 | 26 | 22 | 18 | 14 | 11 | 8 | 24 | 8 | 8 | 4 |

## 4 Experiments

### 4.1 Datasets for Training and Testing

For training, 91 images from Yang et al. [12] and 200 images from the Berkeley Segmentation Dataset were used [13]. We then performed data augmentation on those training images. The images are flipped horizontally, vertically and both horizontally and vertically to make 3 more images for each image. While in the training phase, SET 5 [14] dataset is used to evaluate performance and check if the model is likely to overfit or not. The total number of training images is 1,164 and the total size is 435MB. Color(RGB) images are converted to YCbCr image and only Y-channel is processed. Each training image is split into 32 by 32 patches with stride 16 and 64 patches are used as a mini-batch. For testing, we use SET 5 [14], SET 14 [15], and BSDS100 [13] datasets.

### 4.2 Training Setup

Each CNN is initialized with the method proposed by He et al. [11] and also initialized to 0 for all biases and PReLUs. During training, dropout [16] with p = 0.8 is applied to each output of PReLU layers. Mean Squared Error (MSE) between the estimated output and ground truth is used as a basic loss value and we also add the sum of L2 norms of each CNN's weight (scaled by the factor of 0.0001) to the loss for regularization. We use Adam [17] with an initial learning rate = 0.002 for the optimization algorithm to minimize loss. When the loss does not decrease after 5 epochs of training steps, the learning rate is decreased by a factor of 2 and training is finished if the learning rates goes lower than 0.00002. We also present a compact version of our proposed network (c-DCSCN) as the parameters are shown in Table 1. An example of the results are shown in Figure 3.

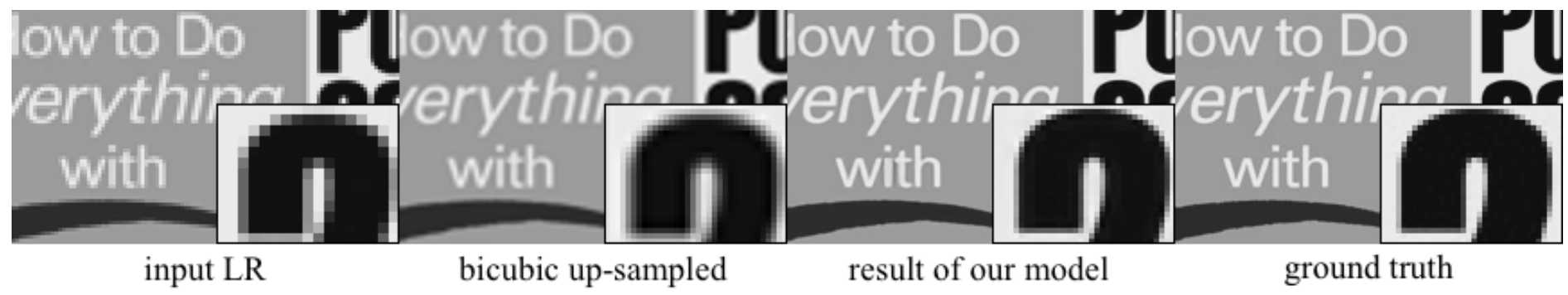

**Fig. 3.** An example of our result of img_013 in set14 [15]

### 4.3 Comparisons with State-of-the-Art Methods

**Comparisons with accuracy** Peak Signal-to-Noise Ratio (PSNR) are used to compare the accuracy of the proposed DCSCN with other Deep Learning-based SR algorithms. Table 2 shows quantitative comparisons for 2x SISR. Red text indicates the best performance and the blue text indicates the second-best. The result shows our proposed algorithm (DCSCN) has either a best or second-best performance for those datasets.

**Table 2.** PSNR/SSIM Comparisons of accuracy with other SR algorithms. (scale = x2)

| Dataset | SRCNN | DRCN | VDSR | RED30 | DCSCN (ours) | c-DCSCN (ours) |
|---|---|---|---|---|---|---|
| Set5 | 36.66/0.9542 | **37.63**/0.9588 | 37.53/0.9587 | **37.66**/**0.9599** | 37.62/**0.9590** | 37.13/0.9569 |
| Set14 | 32.45/0.9063 | **33.04**/0.9118 | 33.03/0.9124 | 32.94/**0.9144** | **33.05**/**0.9126** | 32.71/0.9090 |
| BSD100 | 31.36/0.8879 | 31.85/0.8942 | 31.90/**0.8960** | **31.99**/**0.8974** | **31.91**/0.8956 | 31.60/0.8905 |

**Comparisons with computation complexity** Since each implementation occurs under different platform and libraries, it's not fair to test execution time to compare these methods. Here we calculate the computation complexity of each method instead. Since deep learning computation is usually difficult to parallelize, computation complexity of 1 pixel is used as a good indicator of computation speed. CNN layers are calculated as $size^2$ times input filters times output filters. Bias, ReLU, adding or multiplying layers are calculated as number of filters. When bicubic up-sampling is needed, we calculate it as 16 multiplications and additions. Thus the approximate computation complexity for each method is shown in Table 3. The complexity calculated may slightly differ from true complexity. For example, FSRCNN [9] and RED [7] contain transposed CNN and it needs to pad 0 before processing. However, those differences are much smaller than CNN calculations and therefore are negligible. So they are ignored to create a brief comparison between performance vs. complexity, as shown in Figure 4. We can see our DCSCN has a state-of-the-art reconstruction performance, while the computation complexity is at least 10 times smaller than VDSR [5], RED [7] and DRCN [4].

**Table 3.** Comparisons of approximate computation complexity. (scale = x2) For comparison, we chose f1, f2, f3, n1, n2 = (9,5,5,64,32) for SRCNN and d, s, m = (56,12,4) for FSRCNN

| | SRCNN (9,5,5) | FSRCNN (56,12,4) | DRCN | VDSR | RED30 | DCSCN (ours) | c-DCSCN (ours) |
|---|---|---|---|---|---|---|---|
| CNN layers | 3 | 8 | 20 | 20 | 30 | 11 | 11 |
| CNN filters | 32, 64 | 56, 12 | 256 | 64 | 64 | 32 to 96 | 8 to 32 |
| bias and activation layers | 3, 2 | 7, 7 | 20, 19 | 20, 19 | 0, 36 | 10, 10 | 10, 10 |
| Input image size | x4 | x1 | x4 | x4 | x4 | x1 | x1 |
| **complexity [k]** | **229.5** | **26.2** | **78,083.2** | **2,668.5** | **4,152.8** | **244.1** | **26.1** |

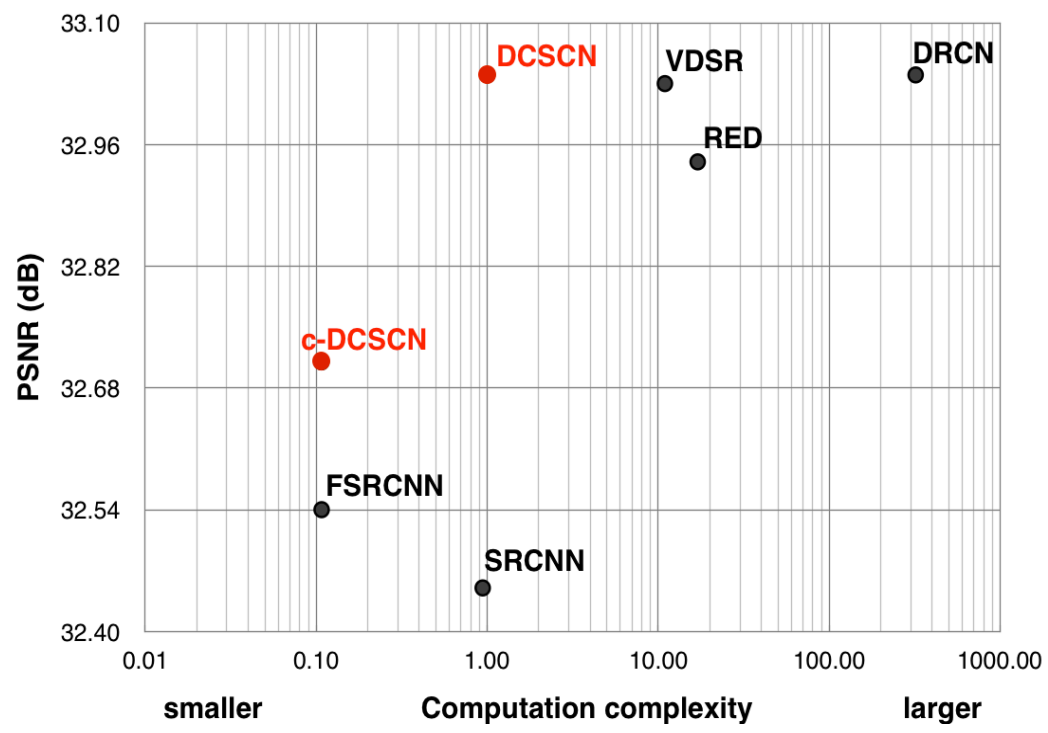

**Fig. 4.** Comparison between reconstruction performance for set14 vs. computation complexity. DCSCN's complexity is taken as 1.00.

## 5 Conclusion and Future works

This paper proposed a fast and accurate Image Super Resolution method based on CNN with skip connection and network in network. In the feature extraction network of our method, the structure is optimized and both local and global features are sent to the reconstruction network by skip connection. In the reconstruction network, network in network architecture is used to obtain a better reconstruction performance with less computation. In addition, the model is designed to be capable of processing original size images. Using these devices, our model can achieve state-of-the-art performance with less computation resources.

Since SISR tasks are now beginning to be used on the network edge (the entry point devices of services like mobile, tablet and IoT devices), building a small but still effective model is rather important. While this model has been proposed through numerous trial and error processes, there should be a better way of tuning the model structure and hyper parameters. Establishment of a method to design suitable model complexity for each problem is needed.

Another noteworthy aspect of this study is the use of the ensemble learning model. Deep Learning itself has a good capacity for complex problems, however, classic ensemble learning tends to lead to good results with less computation, even when there is great diversity within the problem. Also, the ensemble model makes it easier to parallelize for faster computation. Therefore, small sets of Deep-Learning models could be made and combined to work as an ensemble model to fix real and complex problems.